\documentclass[11pt]{article}

\usepackage{graphicx}
\usepackage{subcaption}
\usepackage[utf8]{inputenc}
\usepackage{amsmath, amsfonts}
\usepackage{fullpage}

\begin{document}

\title{Learning to Find Hydrological Corrections}
\author{Lars Arge\thanks{Aarhus University.
    Email: \texttt{large@cs.au.dk}
    }
\and
 Allan Gr\o nlund\thanks{Aarhus University.
   Email: \texttt{jallan@cs.au.dk}
 }
\and
  Svend Christian Svendsen\thanks{Aarhus University.
    Email: \texttt{svendcs@cs.au.dk}
  }
\and
  Jonas Tranberg\thanks{Aarhus University.
    Email: \texttt{stillwalker1234@live.com}
  }  
}

\maketitle

\begin{abstract}
  High resolution Digital Elevation models, such as the (Big) grid terrain model of Denmark with more than 200 billion measurements, is a basic requirement for water flow modelling and flood risk analysis. However, a large number of modifications often need to be made to even very accurate terrain models, such as the Danish model, before they can be used in realistic flow modeling. These modifications include removal of bridges, which otherwise will act as dams in flow modeling, and inclusion of culverts that transport water underneath roads. In fact, the danish model is accompanied by a detailed set of hydrological corrections for the digital elevation model. However, producing these hydrological corrections is a very slow an expensive process, since it is to a large extent done manually and often with local input. This also means that corrections can be of varying quality. In this paper we propose a new algorithmic apporach based on machine learning and convolutional neural networks for automatically detecting hydrological corrections for such large terrain data. Our model is able to detect most hydrological corrections known for the danish model and quite a few more that should have been included in the original list.
\end{abstract}

\section{Introduction}
High resolution Digital Elevation models, such as the grid terrain model of Denmark with more than 200 billion measurements available  as part of the government’s basic data program by the Agency for Data Supply and Efficiency (SFDE) \cite{sfde}, is a basic requirement for several terrain based applications like  water flow modeling and flood risk analysis. However, a large number of modifications often need to be made to even very accurate terrain models, such as the Danish model, before they can be used in realistic flow modeling. These modifications include removal of bridges, which otherwise will act as dams in flow modeling, and inclusion of culverts that transport water underneath roads. For this reason SDFE distribute a detailed set of hydrological corrections for the Denmark model. However, producing these corrections is a very slow and expensive process, since it is to a large extent done manually. This also means that these corrections are of varying quality. Moreover, there are terrain models for many countries that does not come with an official list of  hydrological corrections hindering realistic applications of important hydrological analyses.

The most prominent application of terrain data is probably analyzing the risk of flooding, and the importance of this has only increased by efforts to mitigate the consequences of climate changes. Thus the high costs associated with extreme weather events occurring in densely populated areas has spurred an increased effort into  developing new hydrological
models and methods for analyzing how water flows across terrains in the case of heavy rain and increased sea levels.
Consider a classic simulation of how water flows across a terrain in the event of rain fall. The result of a rain fall may be estimated by first adding some water to all (or subset of) the cells of the terrain model, and then simulating what happens as water flows down hill as follows: In each step water is moved from a one cell to a neighboring cell of lower height, usually the lowest neighboring cell. The simulation considers the cells in order of their height, with the highest cell considered first. In this process each cell may be annotated with the amount of water passing through it. This annotation of the cells is known as \emph{flow accumulation}  \cite{Danner_terrastream, flowacc} and is used  reveal river networks and water ways by extracting the cells with high annotation. The cells that cannot get rid of the water reveal which depressions in the terrain that are flooded \cite{VERDIN19991, Danner_terrastream}.
For such a water flow simulation to produce useful and realistic results, the directions that water flow in the simulation has to (approximately) match how water flows over the surface in real life.
However, a bridge recorded in the digital elevation model breaks this condition, because in real life the water would pass below it, while in the simulation this path is blocked. Hence, obstacles like a bridge that makes the water flow in a wrong direction in the simulation needs to be handled.

\begin{figure*}
  \centering
  \begin{minipage}[t]{0.3\columnwidth}
    \includegraphics[width=1\columnwidth]{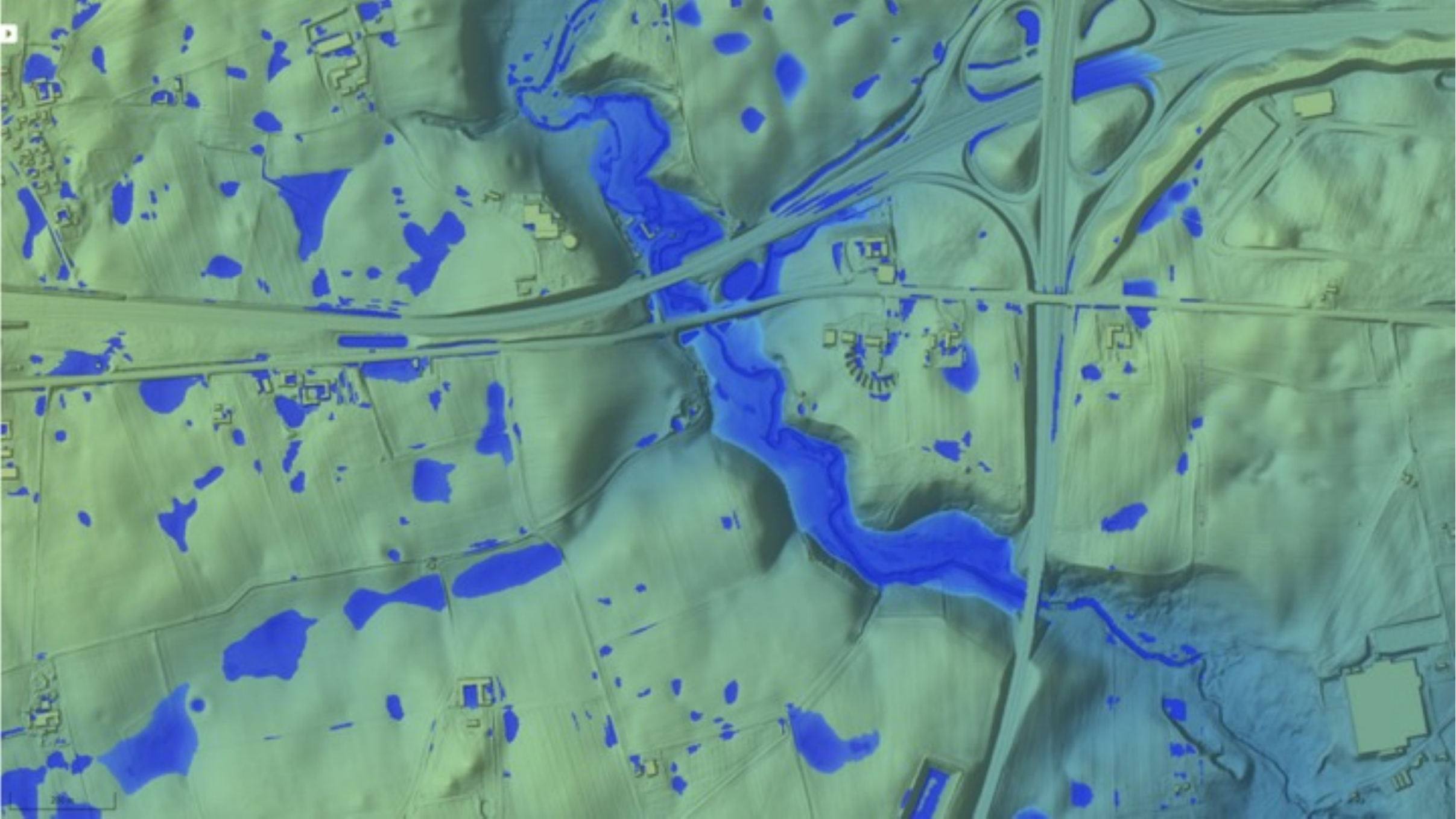}
      \subcaption{Flow Accumulation without hydrological corrections.}
      \end{minipage} $\quad$
  \begin{minipage}[t]{0.3\columnwidth}\centering
    \includegraphics[width=1\columnwidth]{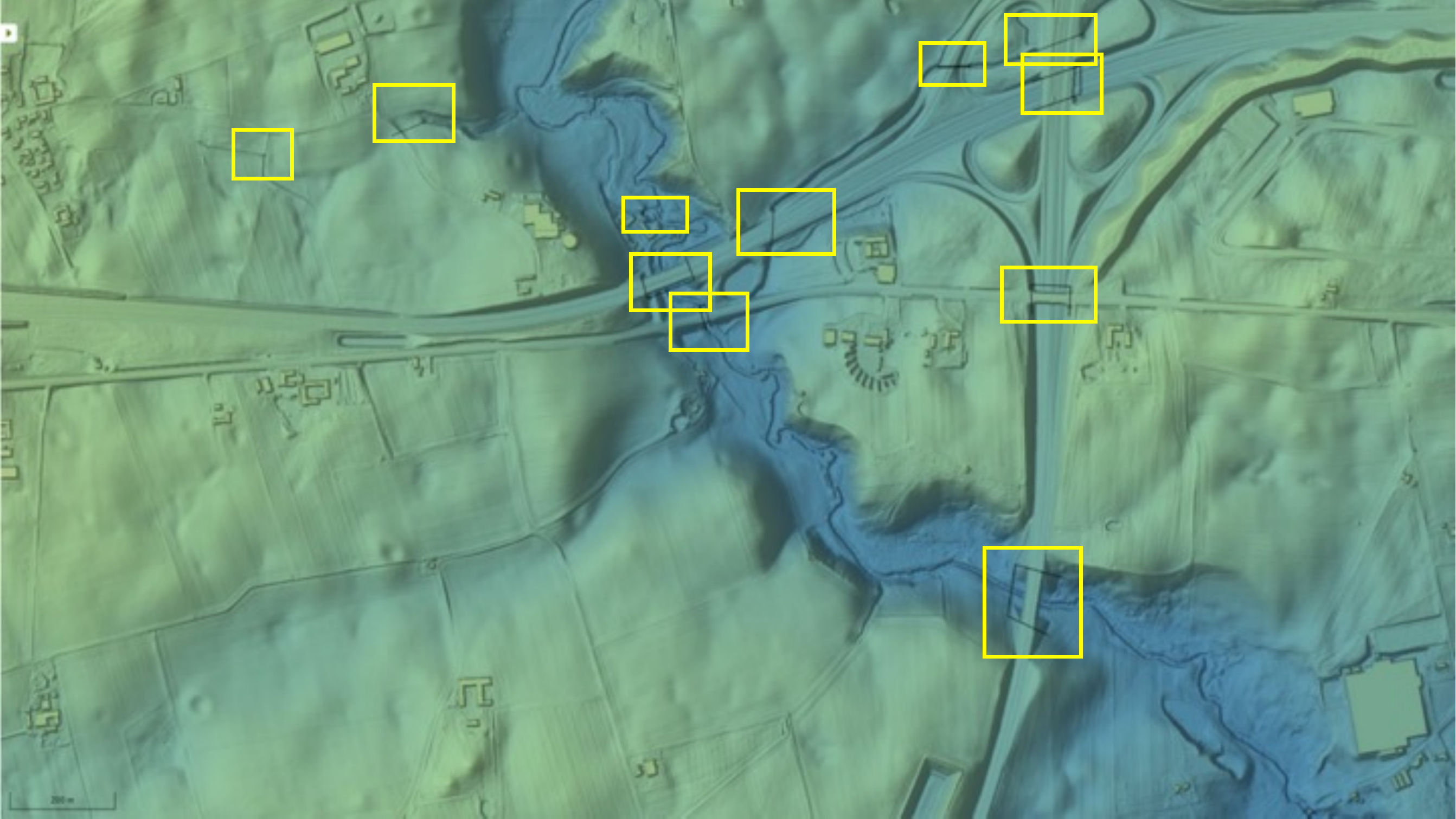}
    \subcaption{Hydrological corrections.\\}
  \end{minipage} $\quad$
  \begin{minipage}[t]{0.3\columnwidth}
    \includegraphics[width=1\columnwidth]{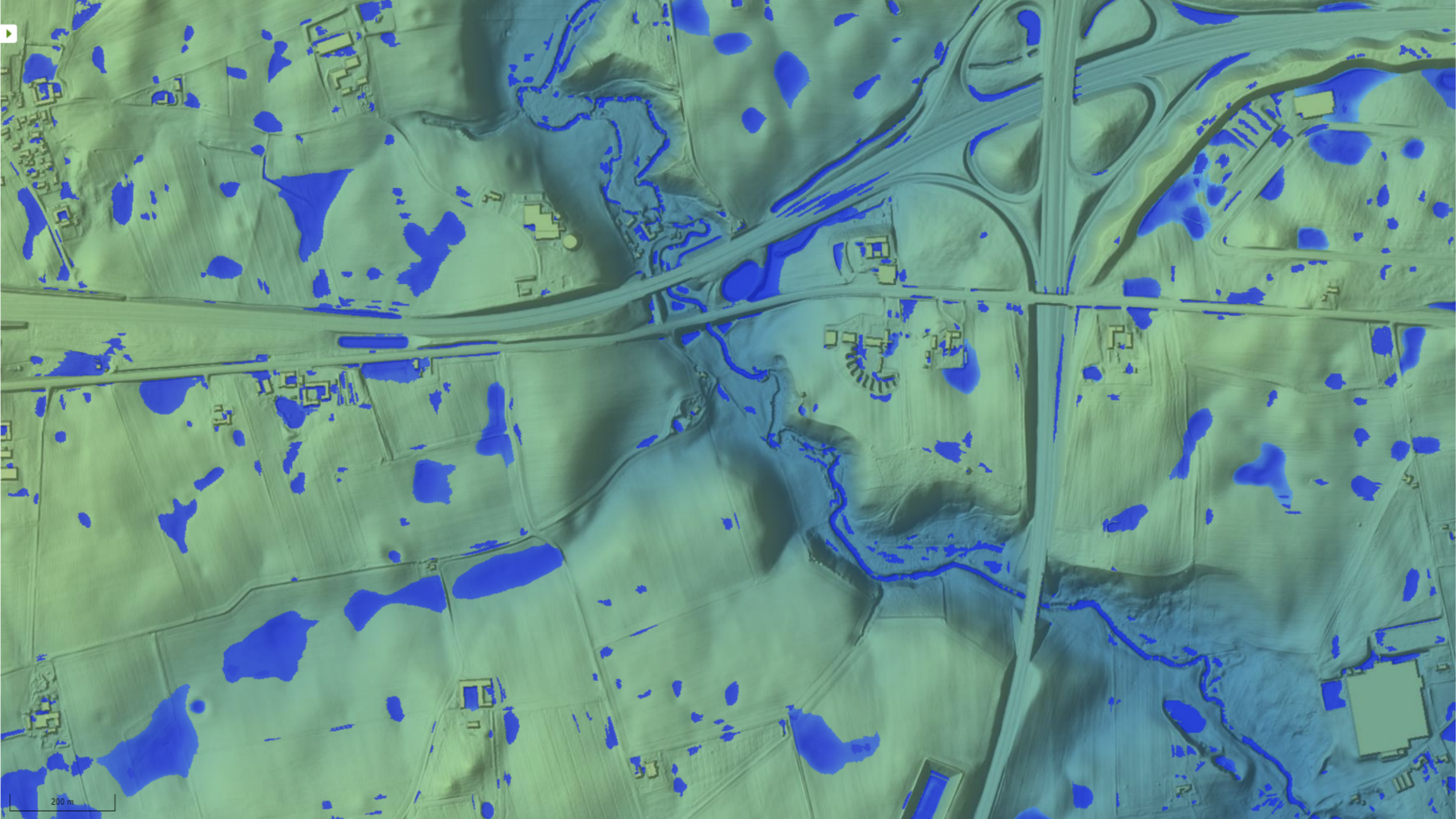}
      \subcaption{Flow Accumulation with hydrological corrections.}
      \end{minipage} $\quad$
  \caption{Visualization of flow accumulation with and without considering hydrological corrections.
    Notice how water accumulates between bridges and on the high way instead of flowing away when hydrological corrections are not considered.}
\label{fig:before_after}
\end{figure*}

We loosely define a hydrological correction as any connected set of cells in the digital elevation model that relative to the surrounding cells has large heights, thus blocking the flow of water in the simulation, where in real life water would actually flow through these cells.
A simple requirement for dealing with the problems created by hydrological corrections is to know where they are. For this reason, lists of hydrological corrections to digital elevation models are sometimes maintained together  with the elevation model, and this list can be used  to update the digital elevation model before any computations are performed. This can for instance be done by cutting the hydrological correction from the elevation model, replacing the heights of the cells comprising the hydrological correction with interpolated heights of the cells of the flow path the hydrological correction blocks.
In Figure \ref{fig:before_after}, a set of hydrological corrections and the results of running flow accumulation with and without considering these hydrological corrections are shown. This figure clearly shows that running analysis that do not consider hydrological corrections returns poor results.
  
Compiling a list of hydrological corrections is usually a manual process. In particular, the list of hydrological corrections for Denmark was made in a manual process where a group of people manually inspected  orthophotos and digital elevation data, focusing primarily on  intersections between road and river networks. Such an approach has several issues. First of all manual labor is slow, expensive, imprecise, and very often inconsistent since deciding whether something is in fact a hydrological correction is hard to pin down exactly. Furthermore,  the manual process needs to be applied again every time the underlying data is changed, which happens continuously. In Denmark the full terrain model is completely updated every five years, each year updating one fifth of the model.  Finally, intersections between road and river networks does not contain all hydrological corrections. For instance, trenches connected with pipes, small streams with small bridges, and tunnels  cannot be found this way. 

\paragraph{Problem Formulation}
The goal is to create an algorithm that automatically locates hydrological corrections in a digital elevation model, and thus automating and improving on the process above. The algorithm takes as input a digital elevation model, along with  other supporting information, such as location of roads and rivers, and the output of water flow algorithms, and outputs a list of potential hydrological corrections including their positions and shapes. 
 We note that we do not really care about very large hydrological corrections (like large bridges) since a list of these is readily available and easy to discover.

\paragraph{Related Work}  Carlson and Danner \cite{carlson:bridge} used feature engineering and machine learning for automated detection of bridge-like objects. The approach they took  was to manually design local feature maps around each cell in an elevation model and then applying the AdaBoost \cite{adaboost} machine learning algorithm  on these features for a cell, trying to predict whether each cell is a part of a hydrological correction. The output of this is then processed by another algorithm that tries to locate the hydrological corrections by grouping areas with many cells predicted as hydrological corrections.
The prediction of whether a given cell is part of hydrological correction or not, is based on five kinds of precomputed features. Carlsen and Danner create  four local feature maps from the digital elevation model:
the first feature is the raw height data, and the next tree features are output of different edge detectors, each based on a 3 x 3 neighborhood around the given cell.
The final feature is a global feature, called a fill map, that is made from a water flow simulation of the entire area in consideration. From each of these feature maps, Carlson and Danner extract 102 features like min, max, mean, avg which totals 510 features per cell.
The data used  in \cite{carlson:bridge} has approximately 6 million cells of  20 feet x 20 feet or 40 feet x 40 feet resolution. To get labeled data, they manually tagged 600 cells of the digital elevation model, 400 negative and 200 positive.

\paragraph{Our Approach:}
Our approach for detecting hydrological corrections along with their position and shape is based on convolutional neural networks. The main ingredient in our algorithm is a convolutional neural network architecture  \cite{lecun_conv} for supervised learning, heavily inspired by convolutional neural networks for image segmentation. Since terrains have high spatial locality, we believe convolutional neural nets that are designed for exactly this situation are the best available tool for the problem, alleviating the need for manually designing features. While the hand designed features designed by Carlson and Danner \cite{carlson:bridge} may to some extent resemble the low level features a convolutional network automatically generate on the same data,  convolutional networks are almost always better at learning useful discriminative features from data with spatial locality than people are at designing them.

The convolutional neural net we employ is designed to  solve the problem on a fixed size tile. More formally, our tile neural network algorithm takes as input a fixed size tile, potentially with several layers of features, and outputs a new  tile of the same size, mapping each cell of the input to the \emph{probability} of whether this cell is a part of a hydrological correction. %
The prediction for each cell is based on the entire tile, allowing the neural network to learn to take advantage of any relevant features within a large area around each cell. Compared to the 3 x 3 cell neighborhood considered in \cite{carlson:bridge}, the neighborhoods we consider are orders of magnitude larger, even when we take into consideration that the cell size in the data we consider is an order of magnitude smaller.
%
The data set of tiles for training the network is initially constructed from  the list of hydrological corrections maintained by SFDE, such that each hydrological correction is contained in at least one tile in the training data. We train a neural network to predict bit maps of the same size as the input tile, where the bits set in the bit map carve out the hydrological corrections contained in the tile.

We solve the full problem of locating all hydrological corrections in an digital elevation model with the tile algorithm as follows: We scan the digital elevation model, splitting it into overlapping fixed size tiles and apply the tile algorithm on these overlapping tiles of the input. The output from the algorithm for these tiles is then combined and used to list all the hydrological corrections and their shapes. 

The data set we use are orders of magnitude larger than the data set considered in \cite{carlson:bridge}, containing approximately 200 hundred billion cells at a resolution of 0.4 meters by 0.4 meters and the list of hydrological corrections from SFDE just shy of 150.000 hydrological corrections.
Hence, the results presented are incomparable to the results achieved by Carlson and Danner \cite{carlson:bridge}. Also, since convolutional nets are considered the state of the art for most image recognition tasks, we have not compared our  approach to theirs.

\paragraph{Our Results}
For the tile problem where the task is to predict the cells that are part of a hydrological correction within the tile, all variations of our algorithm obtain an area under ROC curve (AUC) score between 0.95 and 0.97. The AUC score of an algorithm is equal to the probability it will will rank a randomly chosen cell that is part of a hydrological correction  higher than a randomly chosen cell that is not. We note that the bounding boxes of hydrological corrections in the official list maintained by SFDE has non-negligible variation both in terms of size and position when compared with the the digital elevation model and it is not possible to get perfect accuracy. With this in mind we believe our results for the tile problem are very good. 

For the more general problem of listing all hydrological corrections in an arbitrary sized digital elevation model, we measure how well our algorithm detects the known hydrological corrections. However, we do not have a notion of true negative for this problem, as we do not output where there is not a hydrological correction. This means that we cannot compute an AUC score for this problem. Since an algorithm can propose an excessive amount of hydrological corrections it is important to consider both \emph{precision:} the number of hydrological corrections found divided by the number of hydrological corrections suggested and \emph{recall:} the number of hydrological corrections found divided by the number of hydrological corrections.
Computing the precision and recall statistics is not completely trivial. We need to check if the shape of the hydrological correction output by our  algorithm is \emph{close} to the bounding box of a true hydrological correction. This is complicated by the fact that positions and sizes of ground truth hydrological corrections are noisy, and there may be several hydrological corrections that are close to a proposed hydrological correction.
For our applications recall is more important than precision, and we mainly trade off the two in favor of recall.
All our algorithm variants achieve high recall, but the cost of this is  quite low precision.
This may make our results seems less impressive than we believe they are. There are hydrological corrections in the official list that are almost impossible to detect from the data we have. More importantly, after having analyzed a large number of the false positives, it is clear to us that many of the false positive output by our algorithm are in fact actual hydrological corrections that are just not part of the official list maintained by SFDE.
It is clear to us that the precision of our algorithm is much higher than the tests on the official lists of hydrological corrections suggests, and is in fact a very good algorithm for the problem. Our algorithm has already been included in the commercial product Scalgo Live \cite{scalgolive} where it is being used to detect hydrological corrections in Sweden that does not have an official list of hydrological corrections available.

\paragraph{Paper Outline: }
In Section \ref{sec:data} we describe the data we use in more detail. In Section \ref{sec:segment_theory} we give a short description of the previous work that is the basis for our neural net architecture.  In section \ref{sec:full_alg} we describe the neural net architecture we use for segmenting a tile into the cells that are part of  hydrological corrections and cells that are not. We then describe in detail how we use this neural net algorithm that work for fixed size tiles to detect and output hydrological corrections for the entire digital elevation model.

In Section \ref{sec:experiments} we show the results of our experiments including several actual hydrological corrections output by our algorithm that are not a part of the list of hydrological corrections maintained by SFDE. 

\section{The Data}
\label{sec:data}
In this section we give descriptions of the data and how we construct our initial data set of tiles for the tile algorithm. 
The main data source we consider is the danish digital elevation model which is made  and maintained by The Danish Agency for Data Supply and Efficiency. The digital elevation model is freely available and may be downloaded from \cite{sfde}. The resolution of the model is 0.4 meters, meaning the digital elevation model contains a tiling of Denmark with $0.4m \times 0.4m$ cells each supplied with the height of that cell. This gives a model off approximately 200 billions cells, including parts of ocean which are not relevant for our task.
Besides the digital elevation model, we have extracted road and river network grids from Denmark that we appropriately align with the digital elevation model. Finally, we have made flood computation maps that are also aligned with the digital elevation model. All these we may consider as extra layers of features.


\paragraph{Hydrological Correction Types}
The Danish Agency for Data Supply and Efficiency also makes and maintains a list of hydrological corrections of different types for Denmark.
There are several different kind of hydrological corrections each with different characteristics.
See \cite{sfde_corrections} for the official information on hydrological corrections for the digital elevation model for Denmark including a few examples. The list of hydrological corrections also includes underground pipes that do not leave any marks on the digital elevation models. These are not possible to locate from the data available and we do not consider them. 
The list of these pipes are generated from a separate database that holds the information about  such constructed pipe networks.
The hydrological corrections we consider has two types that are named Horse Shoes and Lines respectively. There are approximately 22.000 Horse Shoe hydrological corrections, and 125.000 Line hydrological corrections in the list for the Denmark model.

\begin{figure}
\centering
\includegraphics[width=0.45\columnwidth]{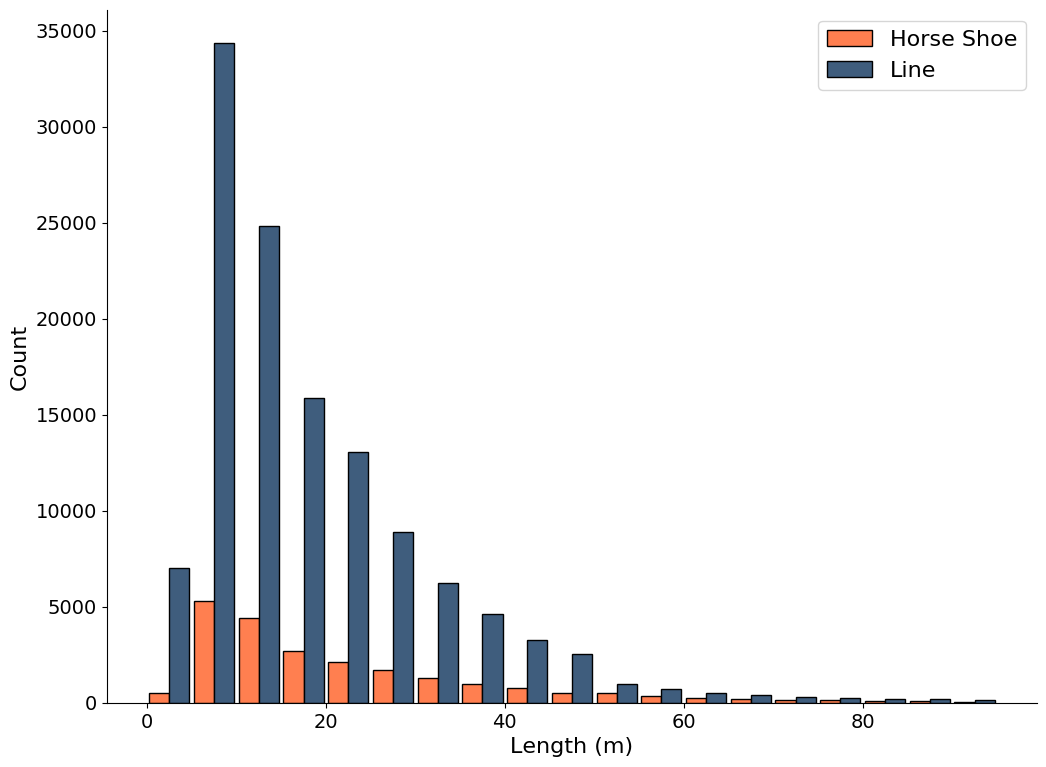}
\caption{Distribution of hydrological corrections lengths.}
\label{fig:corr_size}
\end{figure}

Horse Shoes are hydrological corrections formed by three line segments connected as three sides of a rectangle which resembles the shape of a horse shoe. The Horse Show allow (or disallow) water flowing through an obstacle. A hydrological correction denoted as a line is represented as a single line segment that allow water to flow between the end points. In the data these lines can sometimes be connected into a poly line that lead the water from one end to the other. Such a poly line may be interpreted as one large hydrological correction instead of several small ones but that makes no difference for our purpose, since our algorithm tries to predict all the cells comprising a hydrological correction in the digital elevation model.
We preprocess the hydrological corrections and keep only the Horse Shoe and Line corrections that take up more than one cell.
Inspecting the hydrological corrections in the list compiled by SFDE, it is clear that the size and position relative to the actual corrections one can deduce from the digital elevation model is varying a great deal.
It would of course have been more helpful for us if the true bounding box of every hydrological correction was available, but this is the data that we have.
There also seems to be Line corrections that are sitting on top of completely flat areas, leaving no mark on the digital elevation model, and these essentially acts a noise for our model.
They may actually be  indicating an underground pipe, which we would prefer to remove from the data set, but we cannot deduce it from the information contained in the list of hydrological corrections.
We  note that the size of the individual hydrological corrections vary greatly, from less than one meter to the hydrological correction for the Great Belt Bridge which is close to 7000 meters. The distribution of hydrological correction lengths is shown in Figure \ref{fig:corr_size}.

\section{Segmenting Tiles with Neural Networks}
\label{sec:segment_theory}
Image classification and segmentation algorithms using convolutional neural networks introduced in \cite{lecun_conv} for optical character recognition systems has flourished greatly since the breakthrough paper by Krizhevsky, Sutskever and Hinton \cite{alexnet} that presented a convolutional neural net that outperformed all previous solutions on the famed ImageNet data set by a large margin. Convolutional networks are now the gold standard for several image recognition tasks including image segmentation where the task is to assign the pixels in an image into groups that comprise the relevant different object shown in the image.

As explained in the introduction the goal of our tile algorithm is to segment a tile into the cells that are part of a hydrological correction and the cells that are not.
The very similar and more general problem of predicting pixel level segmentation maps from input images is a well studied problem in the Deep Learning Computer vision field, with a wide range of different models, having different trade-offs. On a high level, the main challenge, when moving from an object detection model (is there a dog in the image), to a pixel level segmentation model (return the pixels that comprise the dog), is the large class imbalance that stem from the fact that most objects only take up a small part of the input image, and the issue of integrating both high level information about the overall presence of an object and low level information about the precise geometric form of the object.

Techniques that tackle the first problem generally fall into two categories. First, there are methods that try to separate the problem into two subproblems: 1) constructing an algorithm that searches the input image for candidate locations for objects and 2) predicting pixel maps from crops of the image at these locations, making the problem significantly more class balanced for the second task \cite{he2017mask}. Secondly, there are methods that try to modify the loss functions to suppress the contribution from pixels that are not part of any object \cite{lin2017focal}.

\begin{figure*}[ht]
  \centering
    \includegraphics[width=\linewidth]{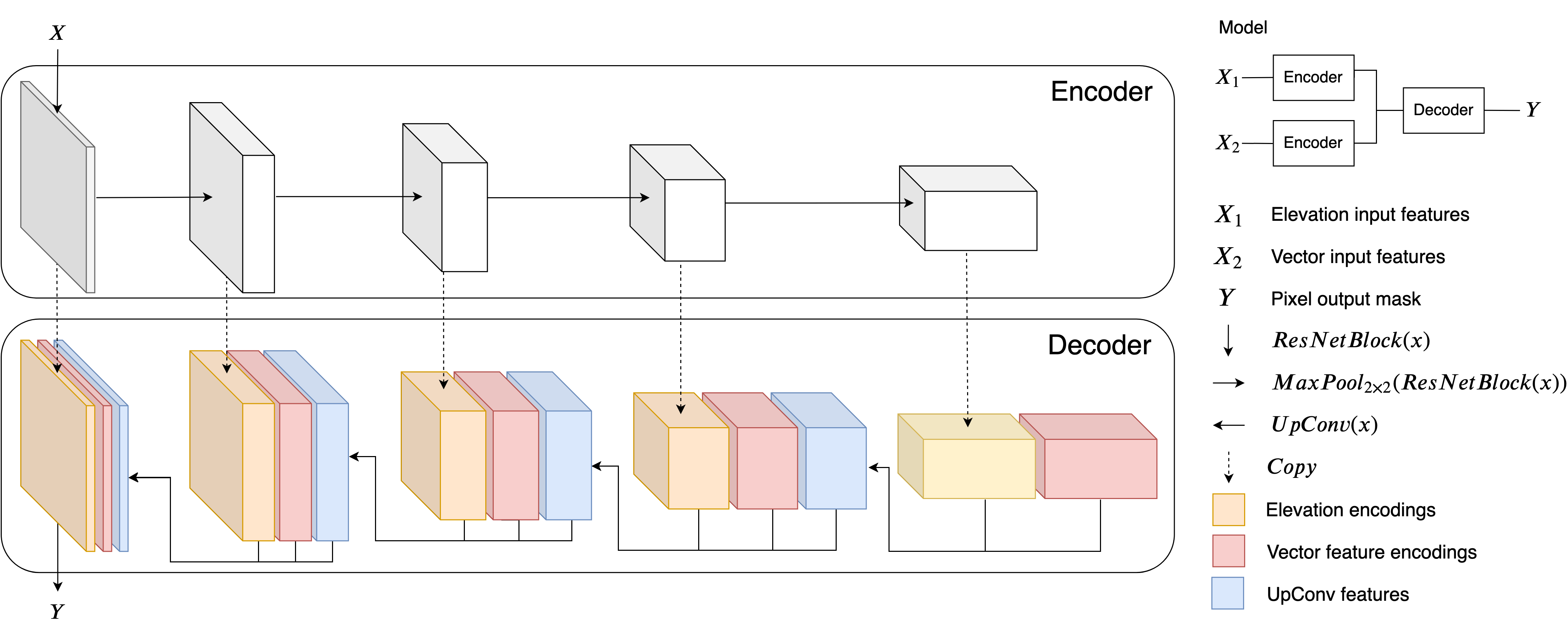}
    \caption{The model. Arrows represent operations, blocks represent data. We use an encoder for each input feature type, as depicted by the \textit{model} subfigure. For example, when using only the elevation map as input feature, the vector feature encodings (red blocks) aren't present. The flow of data is as follows. First the input feature encodings are produced (grey blocks), though a series of operations, each of these decreasing the width and height of the features by a factor 2, while increasing the channel count by a factor 2. These are then used as input to the decoder indicated by the yellow/red blocks. The horizontal $UpConv$ operation then integrate lateral information from the encoder(s) along with more global information from the decoder, each of these \textit{increasing} width and height by a factor 2, while \textit{decreasing} the channel count by a factor 2. Lastly the channel count is reduced to 1 though a single $ResNet$ block and a sigmoid elementwise operation, producing $Y$, representing the probability that a pixel is part of a correction.}
    \label{fig:cor_model} 
\end{figure*}

For the second problem, the integration of both high and low level information about objects, is typically handled through the creation of a feature pyramid. We can separate the feature pyramid network in two processes. First, the \emph{encoder} which increase the channel dimension while decreasing the width and height for increasing layers. Second, the \emph{decoder} which follow up with a decreasing channel dimension and increasing width and height, with concatenated features from the encoder layers. See Figure \ref{fig:cor_model} for a depiction of this process. The hope is that the upsampled features will contain high level information about the presence of objects, while the concatenated channels from previous layers will contain precise information about possible edges of objects. Examples of this is found in U-Net \cite{ronneberger2015u} and Feature Pyramid Networks \cite{lin2017feature}.

Our solution borrows ideas from all of these; We use the U-Net network architecture \cite{ronneberger2015u}, the focal loss to suppress the contribution from low loss pixels from \cite{lin2017focal} and postprocess crops from candidate locations as in Mask R-CNN \cite{he2017mask}.

\section{Complete Algorithm}
\label{sec:full_alg}
In this section we describe our complete algorithm for detecting hydrological corrections in an arbitrary sized region in detail. We start by explaining how we solve the same problem on fixed sizes tiles and  then explain how to use this tile algorithm to analyze an entire region.
Our algorithm works even if the only feature layer we have is the digital elevation model.
Adding more features is straight forward by adding extra layers to the input data aligned with the digital elevation model.

\subsection{Tile Algorithm}
Here we describe our algorithm for locating hydrological corrections in fixed size tiles. This algorithm is a convolutional neural network inspired by convolutional neural networks for image segmentation as described in Section \ref{sec:segment_theory}.

\paragraph{A Data Sets for cell prediction on tiles}
In order to train our neural network to locate the hydrological corrections in a tile we need a data set $D=\{(x_1,y_1),\dots,(x_n,y_n)\}$ which we initially construct as follows.
Each feature tile $x_i$ is a fixed size tile with potentially several feature layers always including a layer with  elevation data from the digital elevation model. The corresponding ground truth element $y_i$ is a tile with one layer of the same size as the feature tile, encoding all the cells within the tile that are a part of a hydrological correction. This encoding is simply a bit mask, where a cell is given the value one if that cell is a part of a hydrological correction, and zero otherwise. We will refer to such a ground truth tile as a label mask.

For a given region of the digital elevation model to learn from we create a data set of tiles as follows. 
For every hydrological correction contained in the region we construct a feature tile and a corresponding label mask with the hydrological correction placed at the center. The feature tile consists of $752 \times 752$ cells from the digital elevation model which we downsample to $376 \times 376$ cells of size $0.8 \times 0.8$ meters. This is done simply to save computation time. Tests have shown that it has no effect on the quality of our algorithm and speeds up our algorithms considerably. The same upsampling is performed on any extra feature layers included. This means that the tiles we consider are squares of approximately $300$ by $300$ meters. We note that the size of our tiles is so large that we can fit all hydrological corrections of interest.
For a given data tile centered around a hydrological correction, we create a the label mask as follows. We start from the all zero tile of the same size as the feature tile and write one in each cell that intersect any of the hydrological corrections in the list of known hydrological corrections for the region, including the hydrological correction at the center of the tile. For the Horse Shoe hydrological corrections this is done by writing a one in each tile cell intersecting the rectangle defined by the Horse Shoe. The Line hydrological corrections are handled the same way by adding a small width to the line segment making into a thin rectangle which is then processed like a Horse Shoe. 


\paragraph{The Loss Function}
\label{sec:loss}
The goal of the training algorithm is to learn a function $f$ that maps the input tiles $x_i$ to the corresponding labels masks $y_i$, such that $f(x_i)\approx y_i$. See Figure \ref{fig:cor_model} for a full specification of the neural net architecture we employ, which as mentioned earlier is heavily inspired by ideas from image segmentation and computer vision. The parameters of the neural net is fit by minimizing the sum of cross entropy loss between between the cells of predicted masks, $f(x)$  and the label mask $y$. We use the focal loss function \cite{lin2017focal} and a special weight map to counter-act the effect of class imbalance in the label masks the algorithm tries to predict. 
The weight map exist to address two concerns that are important for the quality of the final output of our algorithm:
\begin{itemize}
\item On average, only about 1 percent of the cells in a label  mask are part of a hydrological correction and set to one. The rest are zeros.
  If the contribution to the loss from each cell that is part of a hydrological corrections is not higher than the loss associated to the zero valued cells that does not, the learned function becomes heavily biased towards not predicting cells to be part of hydrological corrections.
\item Predicting the label of cells close to a hydrological correction correctly is much more important than getting all the vast amount of easily predicted cells far from a hydrological correction correct. Especially cells at the edge of the hydrological correction are important since down-stream processes vectorize hydrological corrections based on the contour of the prediction maps.
    \item Since hydrological corrections are varying in size, the contribution of large corrections to the loss is  much higher than that of small hydrological corrections if each hydrological correction cell is penalized equally. If this is not handled the learning algorithm puts all emphasis on learning the large hydrological corrections and essentially ignores the small.
\end{itemize}

\begin{figure}
  \centering
  \includegraphics[width=0.8\linewidth]{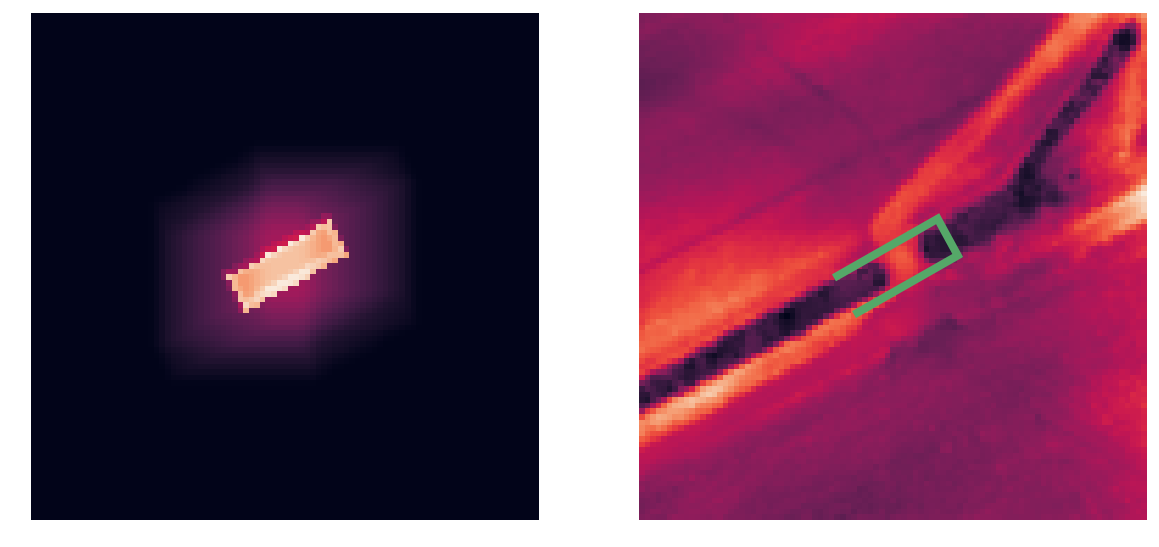}
  \caption{An example of the weight matrix of a single data point. }
  \label{fig:weight_example}
\end{figure}

These concerns give rise to the following definition of the weight-map for scaling the loss to help  the neural network focus on the most important areas of the input and to ensure that the algorithm  learns to detect hydrological corrections of any size.
The weight map $W$ for a data point $(x, y)$ is directly computed from the label mask $y$ as follows. 
$$
W = \frac{1}{\sqrt{wh}} + E + B + L,
$$
where $w$ and $h$ is respectively the width and the height of the tile counted in cells,
\begin{align}
        E &= L * E_\textrm{kernel}, \quad B = E * B_\textrm{kernel}  \\ \nonumber
        L_{ij} & =
        \begin{cases}
            0 &\mbox{if } y_{ij} = 0 \\
            \frac{1}{\mbox{number of cells in correction}} &\mbox{otherwise}
        \end{cases} \\ \nonumber
        E_\textrm{kernel} &=
        \begin{matrix}
            -\frac{1}{8} & -\frac{1}{8} & -\frac{1}{8} \\
            -\frac{1}{8} & 1 & -\frac{1}{8} \\
            -\frac{1}{8} & -\frac{1}{8} & -\frac{1}{8} 
        \end{matrix}, \quad
        B_\textrm{kernel} =
        \begin{matrix}
            \frac{1}{9} & \frac{1}{9} & \frac{1}{9} \\
            \frac{1}{9} & \frac{1}{9} & \frac{1}{9} \\
            \frac{1}{9} & \frac{1}{9} & \frac{1}{9} 
        \end{matrix}
\end{align}
where $*$ is the convolutional operator. Note that cells that are neither part of a hydrological correction nor close to one are weighed as $1/\sqrt{wh}$.
See Figure \ref{fig:weight_example}, for an example of the weight matrix.

The loss for the neural network on a predicted tile is the weighed sum of the losses over the cells of the tile, where the weights are specified by the weight map derived from the label mask. More formally, let $\hat{y}$ be the output mask predicted by the neural network on data point $x$ with label mask $y$, and let $L$ be the weight map induced by $y$. Finally let $\ell$ be the focal loss function from  \cite{lin2017focal}. Then the loss of the network is defined as 
\begin{equation*}
\label{eq:loss}
\sum_{i,j} W_{i,j} \ell \left(\hat{y}_{i,j}, y_{i,j} \right)
\end{equation*}

\paragraph{Training stage}
\label{training-stage}
Our learning algorithm follows the standard practice in image segmentation tasks to boost the number of samples and adding robustness to the learned function, by for each data point $x_i$ considered we first extract a random crop from the feature tile, and then at randomly decide whether to flip the crop on both the horizontal and vertical axis. The same transformation is done on the label mask to predict, and this transformed data point and label mask is then used for training. 


\subsection{Algorithm For General Region}
While we were very successful at recognizing hydrological corrections in the tiles, as we show in Section \ref{sec:experiments}, this does not solve the actual problem posed.
Here we describe how our algorithm finds hydrological corrections in an arbitrary sized region given the algorithm we just described for fixed sized tiles.
First, we cannot just tile the region arbitrarily into fixed size tiles, since that may split hydrological corrections in several pieces, making recognition of them impossible. Such a tiling may also cause an algorithm to report the same hydrological correction several times. Finally, there may be several hydrological corrections in one tile which complicates things further. 

\begin{figure*}
    \includegraphics[width=\linewidth]{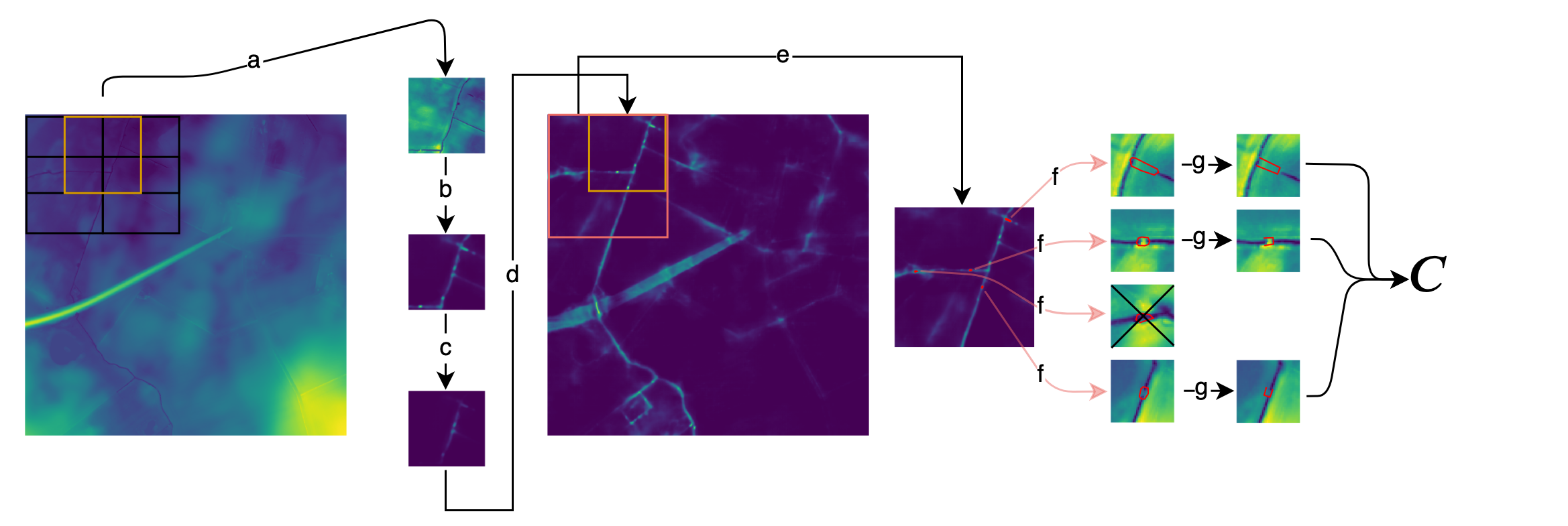}
    \caption{Prediction pipeline:%
      \textbf{a:} Tiles are extracted from the input rectangle in a strided fashion, such that each tile overlap other tiles 50 percent. 
      \textbf{b:} Cell-level probability of hydrological correction membership is predicted for each tile using our trained model for the tile problem defined in Section \ref{training-stage}.
      \textbf{c:} Each tile is then weighed through a monotone window function such that center pixels are weighted $1$ and corner pixels are weighted $0$.
      \textbf{d:} Tiles are added to a probability map of the same size as the input, creating a cell-level probability map of the entire input rectangle. Weighing the tiles with a window function ensure independence of the actual tiling of the region.
      \textbf{e:} As the probability map is filled, a different crop (red rectangle), independent of the tiling in \textbf{a}, is extracted and polygons containing possible corrections are extracted using contours at a fixed threshold.
      \textbf{f:} Each possible correction is evaluated and filtered according to different heuristics. In the above example a correction is filtered because the median probability within the polygon is too low.
      \textbf{g:} Finally, polygons are converted to horse shoe shapes and added to the output.}
    \label{fig:pred_diagram} 
   
\end{figure*}

Without loss of generality we assume the input region to analyze is a, potentially very large, rectangle $M$ including the necessary feature layers that corresponds with the features used in the tile neural network algorithm as described above. The basic idea is to use the tile algorithm on overlapping tiles to generate a new estimated probability map $P$, a rectangle of the same size of $M$, where each cell is associated with the probability of being a part of a hydrological correction, exactly as we did in the tile algorithm. This large map $P$ of probabilities is then processed by searching for areas of high probability and then applying several heuristics to determine if each area found this way is indeed a hydrological correction. Finally, if a hydrological correction has been found, we create a best fit Horse Shoe hydrological correction and add to the set of hydrological corrections that is output at the very end. The full process is visualized and described in more detail in Figure \ref{fig:pred_diagram}.

Formally our algorithm works as follows.
\paragraph{Creating Probability Map}
First we process the input $M$ in a overlap-add fashion, extracting fixed sized crops that fit with the tile algorithm using a stride of $s$ (we sample tiles, $s$ cells apart), creating a set of fixed size tiles that we input into the tile algorithm and save into a list of predicted tiles $X_{i,j}$:
\begin{align*}
    X_{ij}  & = \mbox{nnet}\big(M_{i:(i+2s),j:(j+2s)}\big) \\
    i & =\{0,s, 2s, \dots, h-2s\} \\
    j & =\{0,s, 2s, \dots, w-2s\}
\end{align*}, where nnet is the neural net we created for the tile problem, and $w, h$ is the width and the height of the input rectangle $M$.

The tile predictions are then inserted in prediction map $P$ as follows
$$
P_{i:(i+2s), j:(j+2s)} += H \odot X_{ij}, \quad P\in \mathbb{R}^{h \times w}
$$
where $H=hh^T, h_i = \frac{1}{2} \cos{\bigg(\frac{\pi i}{s}\bigg)}$ is a scaling map that ensures that mainly the predictions for the cells around the center of the tiles are added to the probability map, the further a cell is from the center from the center the more it is scaled down, and $\odot$ is element-wise multiplication.
This finishes the construction of the map of probabilities $P$.

\paragraph{Extracting Hydrological Corrections From Probability Map}
To find the actual hydrological corrections and their shapes we start by creating a contour map on $P$, using a fixed threshold. Each  contour polygon in this contour map represent a possible hydrological correction. We then filter these candidates using the following heuristics designed from manual inspection.

\begin{itemize}
    \item Very small and very large contours are dropped, as most of them are false positives.
    \item Contours with small variance in the elevation data are dropped, as these are mostly false positives. They may also have negligent negative effect on water flow simulations.
    \item The median pixel probability is used as a threshold to control the tradeoff between precision and recall.
\end{itemize}

\paragraph{Outputting Horse Shoes}
Next step is to modify the shape of the contour polygons that the algorithm has decided constitute a hydrological correction.
A given polygon found by our algorithm, describing a hydrological correction, is processed as follows. First we increase the size of the polygon by lowering the probability threshold used in the contour map to gain slightly more context to work with. Then we extract a crop $C$ from the digital elevation model around the polygon.
The cell heights in this tile of  elevation data is then mapped to a probability distribution based on their heights with the lowest values getting the highest probabilities.
We transform the elevation values in the crop $C$ by negating the values, translating them such that the min height is zero and then normalizing by dividing each height value by the sum of heights.

We then sample points from this distribution and fit a Gaussian mixture model with two components to extract the two depressions that the hydrological correction is connecting. This is achieved by picking the mean of the components $\mu_1, \mu_2$ output by by the algorithm as the centers of the two depressions  The line between $\mu_1$ and $\mu_2$ form the skeleton of the connection, while the width is extended in perpendicular direction to the line until it intersect the contour polygon. This give us the resulting horseshoe.

\subsection{Bootstrapping our algorithm}
\label{sec:bootstrap}
As described above, the distribution of zeros and ones in the label masks the neural network for the tile problem must learn to predict, is highly unbalanced. This problem increases significantly when we need to predict hydrological corrections on the entire region considered. In this case the ratio of cells that are part of hydrological corrections is extremely small, much much smaller than in the training data set. The weight map and the focal loss we use to counter this problem help, however with the neural network learned on the initial data set the full algorithm is not able get high recall without predicting relatively many false positives.
To counter this, we analyze the output of the first run of the complete algorithm, and sample new important tiles to learn from for the tile problem. This is achieved by creating tiles centered around false positives, where the predictions of the tile algorithm is close to the decision boundary we use to determine the contour map for the full algorithm. From manual inspection, the false positives far from the decision boundary tend to be actual corrections, revealing incompleteness in the set of manually created corrections. Including these as false positives in our tile algorithm would then make our algorithm worse. With these extra tiles defined we simply restart the training with the new data set, creating a new tile prediction neural network algorithm. We show the results for both in the next section.

\section{Experiments and Results}
\label{sec:experiments}
In this section we describe our experiments.
For training and evaluating our algorithm we use data from the island of Funen, which we have separated along the north-south axis in 2 splits. The training split, which comprise 70 percent of the total area and validation split which comprise 20 percent of the total area\footnote{we set aside the last 10 percent as a test set if we decide to do hyperparameter optimization as future work.}.
Funen has 9000 corrections, split in 5758 Lines and 3299 Horse Shoes

From these splits of Funen, we generate the following data sets:
\begin{description}
\item[bl] Baseline experiment using only the digital elevation model and training only on tiles centered at the hydrological correction.
\item[bs] Bootstrap version of the  baseline experiment (Section \ref{sec:bootstrap}),  with extra tiles centered at locations where the median probability of predicted polygons, using a trained baseline model, is with-in the range $.435 - .45$. We call these extra locations \textit{bootstrapped} locations. 
\item[ff] Like \textbf{bs} but with flash flood features \cite{scalgo:flash}. These features may help the model since flash flood simulations accumulate water at the edge of a correction.
\item[vv] Like \textbf{bs} but with tiles rasterizing road and river vectors as extra layers of features.
  This is expected to help as most intersections between rivers and roads are hydrological corrections.
\item[bs\_wz] Like \textbf{bs} with extra ground truth tiles from the island of Zealand. Zealand has 26651 extra hydrological corrections to consider.
  \item[vv\_wz:] Like \textbf{vv} with extra ground truth tiles from the island of Zealand.  Zealand has 26651 extra hydrological corrections to consider.
\end{description}

The neural network is implemented in Tensorflow, and training on all experiments is done using the ADAM\cite{adam} optimizer with  a learning rate of $.0001$. We use a batch size of $32$ and train on each data set for $50$ epochs. After each epoch, the model is evaluated on the validation set of tiles and the model is saved if the cost has improved.

\begin{table}
  \centering
\begin{tabular}{|l r r r|} 
    \hline
     & AUC & mP & recall \\ [0.5ex] 
    \hline
    bl & 0.969 & 0.2231 & \textbf{0.9126} \\
    \hline
    bs & 0.9692 & 0.3056 & 0.853 \\
    \hline
    ff & 0.9542 & 0.3603 & 0.7845 \\
    \hline
    vv & \textbf{0.977} & \textbf{0.5126} & 0.7482 \\
    \hline
    vv\_wz & 0.9761 & 0.3012 & 0.8498 \\
    \hline
    bs\_wz & 0.9663 & 0.2624 & 0.8626 \\
    \hline
\end{tabular}
\caption{Results for the different data sets. \textbf{AUC} is the area under ROC curve on the validation set of tiles that are generated the same way as the training set.
  \textbf{mP} is the average precision of the centroids of the generated polygons with-in the validation split of Funen, evaluated at a set of thresholds weighted by the change in recall, eg: $mP=\sum_n \big[(Recall_n-Recall_{n-1})/Precision_n \big]$. \textbf{Recall} is the maximal possible recall in the validation region by our full algorithm. That is, how many ground truth corrections are close to a proposed correction, when including all the proposed corrections from prediction pipeline.}
\label{table:results}
\end{table}

\subsection{Results}
We report results for both the tile algorithm and for the algorithm that detects hydrological corrections for an entire input region.
For the tile algorithm, the validation set of tiles we consider is generated the same way as the training set just for a different region. This means tiles centered around a hydrological correction and tiles centered at the bootstrapped locations, except for \textbf{bs}, that only contain tiles centered at hydrological corrections.
This validation score can be evaluated fast, since the area of the tiles is much smaller than the entire region.
The quality of the tile algorithm for predicting which cells in a tile is part of a hydrological correction is evaluated using the area under ROC curve (AUC) score. Reporting pure accuracy is uninformative because of the large class imbalance.

For the full problem of locating hydrological corrections in an entire region, it is not possible to use AUC since the pipeline can propose any number of corrections and "true negatives" are not well defined.  Instead we report \emph{precision}: the ratio between the amount of proposed hydrological corrections close (center-distance$ < 25$ meters) to a true correction (true positive), and all the proposed corrections (true positive + false positive), and \emph{recall}: the ratio between the amount of proposed hydrological corrections close (center-distance$ < 25$ meters) to a true correction (true positive), and the amount of hydrological corrections in the validation region (true positive + false negative). Notice that, in image segmentation tasks, one would usually apply mean intersection over union (mIoU) to determine if a proposed region corresponds to the ground truth shape, but as most of our ground truth hydrological corrections are line shaped, and therefore don't have a well defined area, we use distance to center instead.

The distinction between these two problems is important.
We note that while we are ultimately only interested in the performance on the entire region, it is impractical to train on the entire region by including an excessive amount of extra tiles without any hydrological corrections. That would also add significantly to the label imbalance problem discussed in Section \ref{sec:full_alg}.

\begin{figure}
  \centering
    \includegraphics[width=0.8\linewidth]{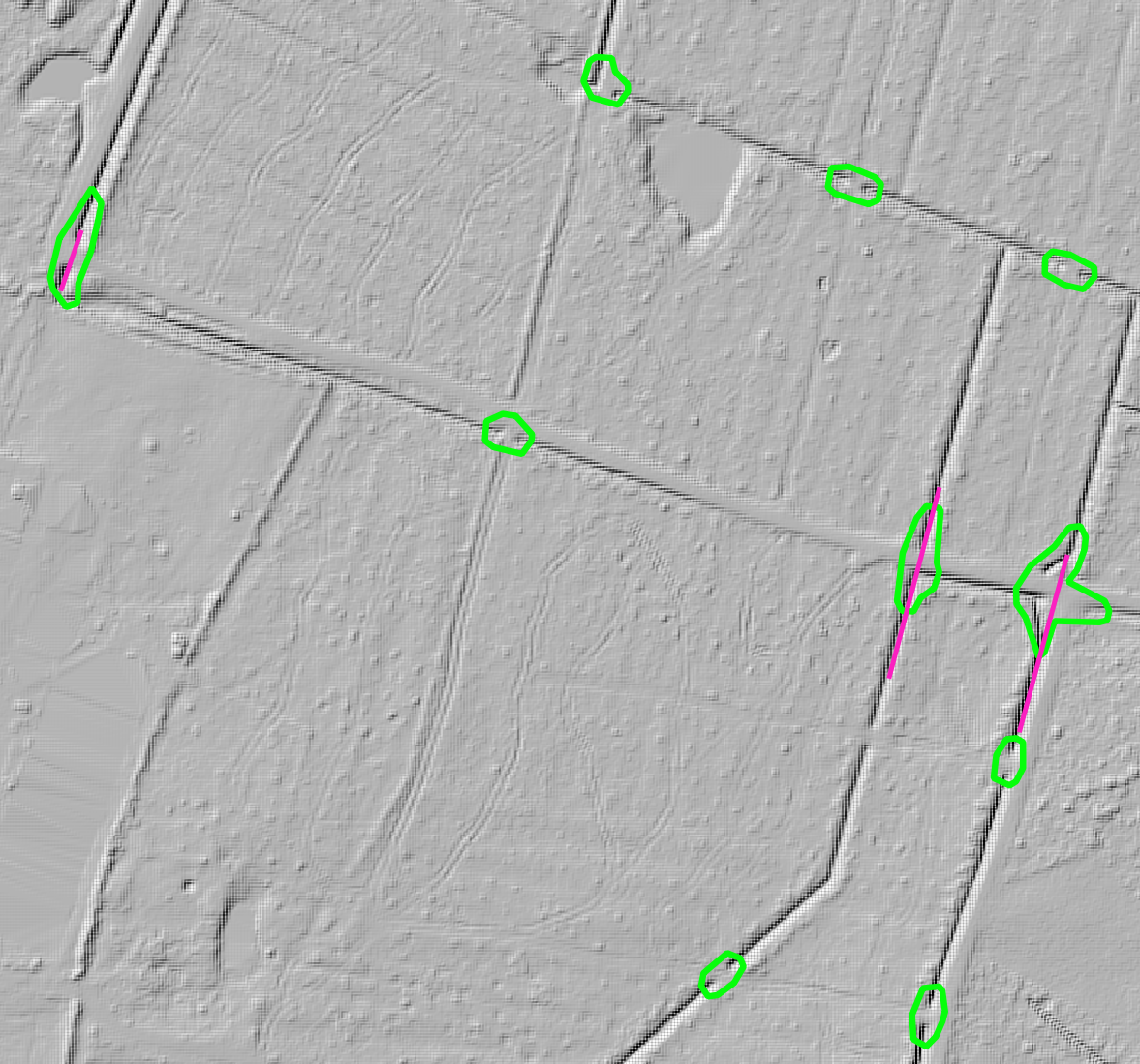}
    \caption{An example actual corrections not in the ground truth set. The purple lines are hydrological corrections from the official list of hydrological corrections and the green polygons are hydrological corrections proposed by our algorithm.}
    \label{fig:false_positives}  
\end{figure}

Detecting hydrological corrections on an entire region is a significantly harder problem than predicting pixel probabilities on tiles, since hydrological corrections are very rare and the distribution of non-correction locations is suspected to be complex. As mentioned earlier, we try to handle this problem, by including non-correction tiles in the training and validation set, whose centers have median probability close to the decision boundary. Perhaps surprisingly, we do not sample false positive locations which have median probability above $.45$, since,  manual inspection reveal that many such false positive locations, are in fact true positives. See Figure \ref{fig:false_positives} for an example with several false positives that are actually true positives. Including these as non-correction tiles in the bootstrapping, would only degrade performance. Predicting hydrological corrections on the  region of Funen takes between 30 minutes and an hour, depending on the number of predicted hydrological corrections, on a dual NVIDIA 1080ti GPU's and a Intel Xeon E5-1650 CPU. Training the neural network for the tile algorithm takes approximately six hours when only considering tiles from Funen.

\begin{figure}
  \centering
    \includegraphics[width=0.4\linewidth]{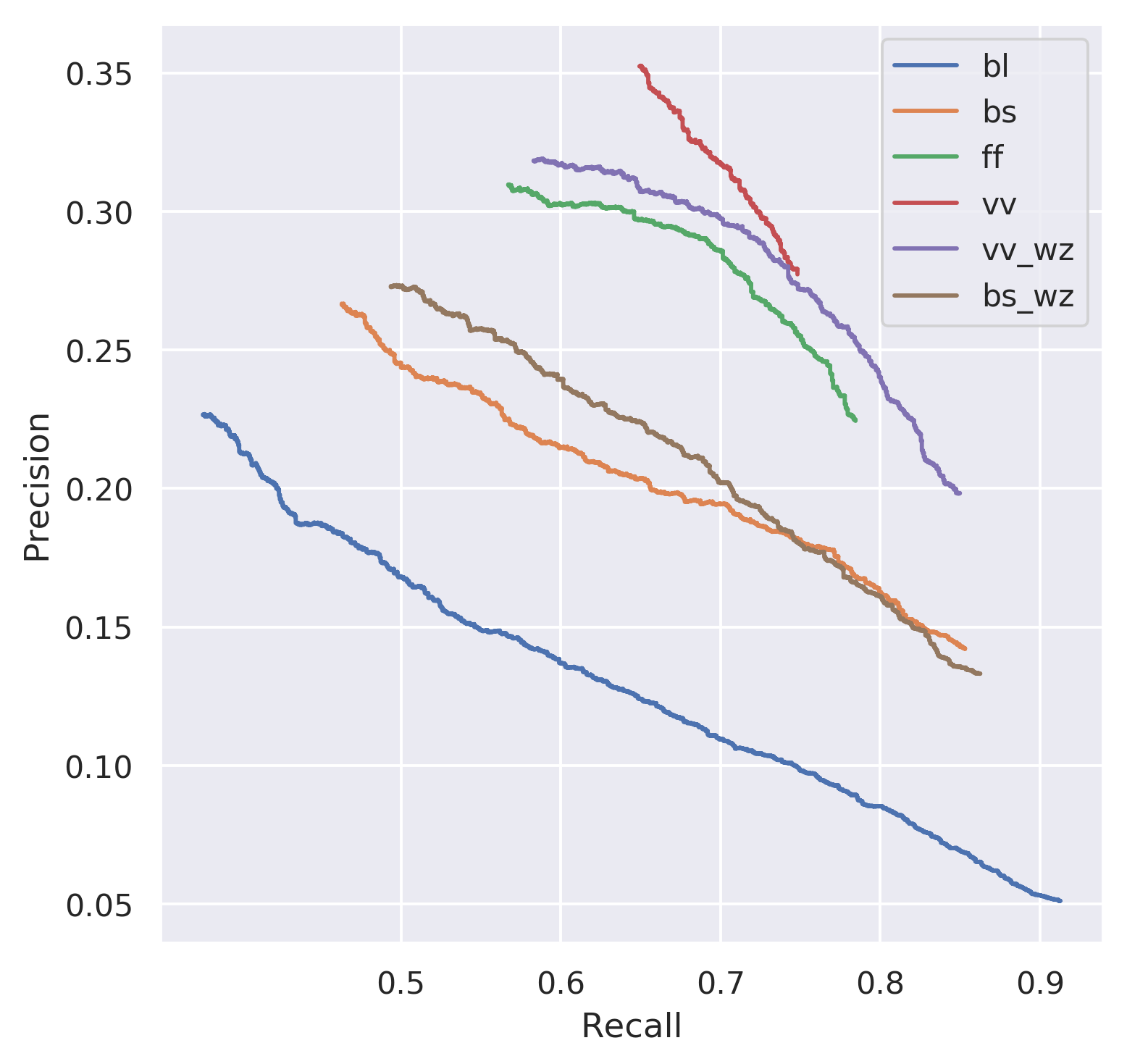}
    \caption{Precision/Recall curves. The precision and recall trade off is determined by the median pixel probability with-in each proposed polygon.}
  \label{fig:precision_recall}
\end{figure}

The discrepancy between the problem of predicting cells in the validation tiles and predicting shapes of corrections on the entire region is shown in Table \ref{table:results}, where all experiments show good performance on the validation tiles; all with-in $0.95-0.97$ AUC. But we also see, that the performance on the validation tiles, does not necessarily translate to good performance on the entire region. For example, the \textbf{vv} experiment has the best AUC ($0.977$), but, when using this model in the prediction pipeline, it proposes too few hydrological corrections, resulting in lowest recall of all experiments. On the other hand, the baseline experiment actually have the best recall of all the experiments, but not very good precision. To gain better understanding of this relationship we show the different trade-off curves in Figure \ref{fig:precision_recall} for the full algorithm based on the neural network trained on the different data sets.

Inspecting figure \ref{fig:precision_recall}, it not clear that any model is ultimately better, since they all trade maximal recall for precision. One exception is vv\_wz which achieve the same maximal recall as both \textbf{bs} experiments while maintaining a much better precision. Its also clear that including the bootstrapped locations improve precision significantly.

\section{Conclusion and Future Work}
In this paper we have described a new approach for detecting hydrological corrections that automates and improves the existing manual process. Our algorithms find almost all known hydrological corrections, and finds many more that should have been included in the list. The many missing hydrological corrections from the  list maintained by SFDE is a problem both in terms of reporting how well an algorithm actually works but it is also a significant issue because the labels the algorithm  learn from become noisy. As mentioned above, another issue with the official data is that the exact position and shape of the hydrological corrections in the list vary greatly when compared with the underlying digital elevation model. This makes both tile problem and the full problem harder. From our experiments our algorithm for the tile problem seems to be fairly robust to this problem.
An industrial strength version of our algorithm have been implemented and incorporated into the commercial product of Scalgo Live\cite{scalgolive}. This algorithm uses only digital elevation model which is often the only data  available. Our algorithm is currently only used for Sweden that does not have any official list of known hydrological corrections. To help our algorithm we have acquired 1500 hydrological corrections from three different Swedish cities and added to the hydrological corrections from Denmark to train on. We use the bootstrapped version of our algorithm which gives the best tradeoff between precision and recall. The Swedish model has a resolution of $2m \times 2m$ and it took 3 days on a standard, single GPU work station to run our full algorithm on the entire country.

There are several avenues to explore for further improvement of our algorithm mainly to improve precision. We believe the most promising strategy is to improve  the quality of the list of hydrological corrections since this will help all parts of the process, from the learning algorithm, to reporting more truthful precision and recall statistics. The latter is very important since it is hard to improve on our algorithm when the measure we use to compare algorithms is noisy. For this reason we are currently running a project where different experts and end users in the field are shown the false positives output by our algorithm and then has to decide by manual inspection whether the false positive is actually a hydrological correction or not. 

\bibliographystyle{plain}
\bibliography{main}
\end{document}